\documentclass[a4paper]{article}
\usepackage[a4paper]{geometry}

\usepackage[utf8]{inputenc}
\usepackage[T1]{fontenc}

\usepackage{natbib}
\usepackage{orcidlink}

\usepackage{authblk,url,booktabs,nicefrac,microtype,xcolor,tikz,xr}
\usepackage{amsmath,amsfonts,amssymb,amsthm,bm,mathtools,pifont}
\usepackage{mdframed}

\usepackage{hyperref}
\usepackage{cleveref}

\usepackage[toc,page,header]{appendix}
\usepackage{minitoc}

\newcommand{\defeq}{\vcentcolon=}
\DeclareMathOperator*{\argmin}{arg\,min}

\newcommand{\T}{\top}

\newcommand{\kl}{\mathtt{KL}}
\newcommand{\tr}{\mathtt{tr}}
\newcommand{\dx}{\mathtt{d}\bm{x}}

\newcommand{\dy}{\mathtt{d}\bm{y}}
\newcommand{\dbary}{\mathtt{d}\bar{\bm{y}}}

\newcommand{\calC}{\mathcal{C}}
\newcommand{\calD}{\mathcal{D}}
\newcommand{\calM}{\mathcal{M}}
\newcommand{\calT}{\mathcal{T}}
\newcommand{\calX}{\mathcal{X}}
\newcommand{\calY}{\mathcal{Y}}
\newcommand{\calZ}{\mathcal{Z}}
\newcommand{\calU}{\mathcal{U}}
\renewcommand{\Re}{\mathbb{R}}

\usepackage{xspace}
\makeatletter
\DeclareRobustCommand\onedot{\futurelet\@let@token\bmv@onedotaux}
\def\bmv@onedotaux{\ifx\@let@token.\else.\null\fi\xspace}
\def\eg{\emph{e.g}\onedot} 
\def\ie{\emph{i.e}\onedot} 
 
\def\etc{\emph{etc}\onedot} 
\def\wrt{w.r.t\onedot}

\def\iid{i.i.d\onedot}
\def\aka{a.k.a\onedot}

\makeatother

\graphicspath{{./}{./figures/}}
\newtheorem{theorem}{Theorem}
\newtheorem{lemma}[theorem]{Lemma}
\newtheorem{corollary}[theorem]{Corollary}
\newtheorem{proposition}[theorem]{Proposition}
\newtheorem{definition}[theorem]{Definition}
\newtheorem{remark}{Remark}[theorem]
\newmdtheoremenv[
  outerlinewidth=1,
  roundcorner=5pt,
  leftmargin=3,
  rightmargin=3,
  backgroundcolor=yellow!15,
  innertopmargin=3pt,
  splittopskip=3pt,
  ntheorem=true,
]{mydefinition}[theorem]{Definition}

\newcommand{\xdasharrow}[2][->]{
\tikz[baseline=-\the\dimexpr\fontdimen22\textfont2\relax]{
\node[anchor=south,font=\scriptsize, inner ysep=1.5pt,outer xsep=2.2pt](x){#2};
\draw[shorten <=3.4pt,shorten >=3.4pt,dashed,#1](x.south west)--(x.south east);
}}

\newmdenv[innerlinewidth=0.5pt,roundcorner=1pt,backgroundcolor=yellow!10,
linecolor=gray,
innerleftmargin=6pt,innerrightmargin=6pt,innertopmargin=6pt,innerbottommargin=6pt]{notice}

\hypersetup{colorlinks=false}
\title{Intrinsic Universal Measurements of Non-linear Embeddings}
\author{Ke Sun\,\orcidlink{0000-0001-6263-7355}}
\affil{CSIRO Data61, Sydney, NSW, Australia\\The Australian National University, Canberra, ACT, Australia\\Email:~\texttt{sunk@ieee.org}\\\url{http://courbure.com}}
\date{}
\hypersetup{colorlinks=true, citecolor=blue}

\makeatletter
\begin{document}
\maketitle

\doparttoc\faketableofcontents

\begin{abstract}
A basic problem in machine learning is to find a mapping $f$ from a low
dimensional latent space $\calY$ to a high dimensional observation space
$\calX$. Modern tools such as deep neural networks are capable to represent
general non-linear mappings. A learner can easily find a mapping which perfectly
fits all the observations. However, such a mapping is often not considered as good,
because it is not simple enough and can overfit. How to define simplicity? We try to make
a formal definition on the amount of information imposed by a non-linear mapping $f$.
Intuitively, we measure the local discrepancy between the pullback
geometry and the intrinsic geometry of the latent space. Our definition is based
on information geometry and is independent of the empirical observations, nor specific
parameterizations. We prove its basic properties and discuss relationships with
related machine learning methods.
\end{abstract}


\vspace{5em}
\begin{notice}
Cite as:\\
Ke Sun. Local measurements of nonlinear embeddings with information geometry.
In Frank Nielsen and Arni S.R. Srinivasa Rao and C.R. Rao, editors,
\emph{Handbook of Statistics}, volume 46, pages~257--281, Elsevier, 2022.
DOI: \url{https://doi.org/10.1016/bs.host.2022.03.008}
\end{notice}

\newpage

\section{Introduction}\label{sec:intro}

\begin{figure}[b]
\centering
\includegraphics[width=.5\textwidth]{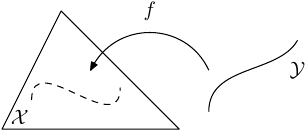}
\caption{The basic subjects in this paper: a latent Euclidean space $\calY$;
an observation space $\calX$ (a Riemannian manifold);
a differentiable mapping $f:\,\calY\to\calX$ from $\calY$ to $\calX$.}\label{fig:f}
\end{figure}

In statistical machine learning, one is often interested to derive a non-linear
mapping $f$, from a latent space $\calY$ to an observable space $\calX$
(as shown in \cref{fig:f}),
so that meaningful latent representations can be learned for the observed data.
Similar problems widely appears in dimensionality reduction (\aka manifold learning),
non-linear regression, and deep representation learning.

By convention in machine learning,
we refer the inverse of $f$ which we denote as $g$, rather than $f$ itself, as
an ``embedding'' that is a mapping from $\calX$ to $\calY$. The latent space
$\calY$ is referred to as the embedding space. In Riemannian geometry,
an embedding is from the low dimensional $\calY$ to the high dimensional
$\calX$, but not the other way round. This is only a matter of terminology.

How to measure and learn an embedding?
Consider a non-linear mapping $f_{\bm\theta}:\,\calY\to\calX$ parameterized by a
neural network with parameters $\bm\theta$ (the weights and bias terms).
The discrepancy of $f_{\bm\theta}$ with respect to (\wrt) a given observable
$\bm{x}\in\calX$ and a latent $\bm{y}\in\calY$ can be measured by
some distance between $f_{\bm\theta}(\bm{y})$ and $\bm{x}$ denoted by
$D(f_{\bm\theta}(\bm{y})\,:\,\bm{x})$.
This is illustrated in the following diagram:
\begin{equation*}
\begin{array}{ccccl}
\bm{y} & \xrightarrow{\bm\theta} & f_{\bm\theta}(\bm{y}) & \searrow & \\
&&&& D( f_{\bm\theta}(\bm{y})\,:\,\bm{x} ) \quad \text{(data geometry)}\\
&& \bm{x} & \nearrow &\\
\end{array}
\end{equation*}
As both $f_{\bm\theta}(\bm{y})$ and $\bm{x}$ are points on $\calX$, such
measurements of $f$ is based on the data geometry, which is the geometry of
$\calX$ equipped with fundamental quantities such as the distance.
Then, one can minimize the empirical average of
$D(f_{\bm\theta}(\bm{y})\,:\,\bm{x} )$ \wrt a set of samples to learn
the neural network $f_{\bm\theta}$.
Similarly, an embedding $g_{\bm\theta}:\,\calX\to\calY$
can be measured by first mapping $\bm{x}$ to the latent space, and then
measuring the distance against $\bm{y}$. The following diagram
shows this paradigm based on the geometry of $\calY$, the embedding geometry.
\begin{equation*}
\begin{array}{cccll}
&& \bm{y} & \searrow & \\
&&&& D( \bm{y}\,:\,g_{\bm\theta}(\bm{x}) ) \quad \text{(embedding geometry)}\\
\bm{x} & \xrightarrow{\bm\theta} & g_{\bm\theta}(\bm{x}) & \nearrow &\\
\end{array}
\end{equation*}
A typical example of these paradigms is autoencoder networks~\citep{kingma14,rifai11},
where $g_{\bm\theta}$ and $f_{\bm\theta}$ are called the ``encoder''
and the ``decoder'', respectively.

As a convention of notations, we use Greek letters such as ``$\alpha$'' and ``$\beta$'' to denote scalars,
bold lowercase letters such as ``$\bm{x}$'' and ``$\bm{y}$'' for vectors,
regular capital letters such as ``$A$'' and ``$W$'' for matrices,
and calligraphic letter such as ``$\calX$'' and ``$\calY$'' for manifolds
and their related measurements, with exceptions.
We reserve the symbol ``$D$'' for a general distance-like measure,
and the symbol ``$\calD$'' for our proposed discrepancy measure.
The mappings are denoted by $f_{\bm\theta}$ and $g_{\bm\theta}$,
where the subscript $\bm\theta$ (parameters of the associated mapping)
can be omitted.

Another type of embedding is based on information theoretical measurements,
which fits in the following paradigm:
\begin{equation*}
\begin{array}{ccccc}
\bm{y} & \xrightarrow{\text{embedding geometry}} & p & \searrow &\\
    &&&& D(p\,:\,q) \quad\text{(information geometry)}\\
\bm{x} & \xrightarrow{\text{data geometry}}     & q & \nearrow & \\
\end{array}
\end{equation*}
Two probability distributions
$p$ and $q$ are computed for $\calX$ and $\calY$, respectively. Then the
quality of embedding is measured by some ``distance'' $D(p\,:\,q)$ between these two
probability distributions, based on the geometry of the space of probability
distributions, \aka information geometry~\citep{amari16}.
A typical example is stochastic neighbourhood embedding~\citep{hinton03},
where $p=(p_{ij})_{n\times{n}}$ and $q=(q_{ij})_{n\times{n}}$ denote pairwise
proximity \wrt $n$ samples on the two manifolds $\calX$ and $\calY$,
respectively, and $D(p:q)$ is chosen as the Kullback-Leibler (KL) divergence.

In summary, the following three geometries are involved to measure $f:\,\calX\to\calY$:
\begin{description}
    \item[Data Geometry]
        Geometric measurements in a usually high dimensional observation space denoted as $\calX$,
        or the geometry of the input feature space;
    \item[Embedding Geometry]
        Intrinsic geometry of the embedding space $\calY$, which is usually
        assumed to have a simple well-defined geometric structure, such as the
        Euclidean space;
    \item[Information Geometry]
        Geometry of the space of probability distributions.
        It also gives an intrinsic geometry of parametric probabilistic models.
\end{description}
It is a fundamental and mostly \emph{unsolved} problem on how to properly define
these geometries, and how these geometries interact and affect learning.

Following the above information geometric approach, this work tackles the basic
problem of defining an \emph{intrinsic complexity} of the mapping $f:\,\calY\to\calX$.
Our construction is based on the following intuitions. For any
given $\bm{y}\in\calY$, we can define a local probability distribution $p(\bm{y})$
(a ``soft neighbourhood'' of $\bm{y}$) based on how the image of $\calY$
under the mapping $f$ is embedded inside $\calX$.
In geometric terms, $p(\bm{y})$ is defined based on a submanifold $f(\calY)$
of the ambient space $\calX$, which is often referred to as the ``data manifold''.
On the other hand, we can build another not necessarily normalized
measure $s(\bm{y})$ around $\bm{y}$ based on the intrinsic structure of $\calY$.
For a low complexity $f$, $p$ and $s$ should be similar, meaning that
the mapping $f$ does not impose additional information;
$p$ and $s$ are dissimilar for a high complexity $f$.
Hence, we measure the complexity, or imposed information,
of $f$ locally around $\bm{y}\in\calY$ through measuring the discrepancy of $p$ and $s$.

We made the following contributions:
\begin{itemize}
\item A formal definition of $\alpha$-discrepancy that is a one-parameter
      family of discrepancy measurements of a differentiable mapping $f:\calY\to\calX$;
\item Proof of its invariance and other basic properties;
\item Discussion on its computational feasibility and practical estimation methods;
\item Connections with existing machine learning methods including
      neighbourhood embeddings and autoencoders.
\end{itemize}

In the rest of this chapter, we first review the $\alpha$-divergence, based on
which our core concepts are defined (\cref{sec:alpha}). We formally define the
local and global $\alpha$-discrepancy of an embedding and
show its basic properties and closed-form expressions (\cref{sec:embed}).
Then, we demonstrate that the $\alpha$-discrepancy is
computationally feasible and give practical estimation methods
(\cref{sec:empirical}). We present how it connects with existing techniques such as
the neighbour embeddings and related deep learning methods (\cref{sec:connections}).
We discuss possible extensions and concludes (\cref{sec:con}).
Proofs of the statements are provided in the end of the chapter.

\section{$\alpha$-Divergence and auto-normalizing}\label{sec:alpha}

Recall that in our construction, we need a ``distance'' between two local positive measures
on the manifold $\calY$: one is associated with a neighbourhood on the data manifold
$f(\calY)$ around a reference point $\bm{y}\in\calY$; the other means a neighbourhood based
on the underlying geometry of $\calY$ by prior assumption (that is independent of $f$).
Therefore we fall back to defining information theoretical distances
in the space of positive measures.

Information geometry~\citep{amari16} is a discipline where information
theoretic quantities are endowed with a geometric background,
so that one can study the ``essence of information'' in an intuitive way.
It has broad connections with statistical machine
learning~\citep{amari95,lebanon02,sun14,yang14,amari16}. We only need one of
the many useful concepts from this broad area and refer the reader to
\citep{amari16,ignielsen} for proper introductions.

The $\alpha$-divergence~\citep{amari16}
is a uni-parametric family of \emph{information divergence} (non-negative dissimilarity
that is asymmetric and does not satisfy the triangle inequality).
Given two finite positive measures $\tilde{p}$ and $\tilde{q}$
(that are not necessarily normalized)
defined on the same support $\calY$, the $\alpha$-divergence gauges
their ``distance'' by
\begin{equation}\label{eq:alphadiv}
D_\alpha(\tilde{p}\,:\,\tilde{q})
\defeq
\frac{1}{\alpha(1-\alpha)} \int \left[
  \alpha \tilde{p}(\bm{y})
+ (1-\alpha) \tilde{q}(\bm{y})
- \tilde{p}^{\alpha}(\bm{y}) \tilde{q}^{1-\alpha}(\bm{y})
\right] \dy
\end{equation}
for $\alpha\in\Re\setminus\{0,1\}$. If $p$ and $q$ are probability distributions, then
\cref{eq:alphadiv} is simplified to
\begin{equation}\label{eq:alphadivnormalized}
D_\alpha(p\,:\,q)
=
\frac{1}{\alpha(1-\alpha)} \left[
1 -\int p^{\alpha}(\bm{y}) q^{1-\alpha}(\bm{y}) \dy
\right].
\end{equation}

It is easy to show from L'H\^{o}spital's rule that
\begin{equation*}
\lim_{\alpha\to1}D_{\alpha}(\tilde{p}\,:\tilde{q})
=
\kl(\tilde{p}\,:\,\tilde{q})
=
\int \left[
    \tilde{p}(\bm{y}) \log\frac{\tilde{p}(\bm{y})}{\tilde{q}(\bm{y})}
    - \tilde{p}(\bm{y}) + \tilde{q}(\bm{y})
\right]\dy
\end{equation*}
is the (generalized) KL divergence, and similarly,
$\lim_{\alpha\to0}D_{\alpha}(\tilde{p}\,:\,\tilde{q})=\kl(\tilde{q}\,:\,\tilde{p})$
is the reverse KL divergence.
Therefore the definition of the $\alpha$-divergence is naturally extended to $\alpha\in\Re$,
encompassing KL, reverse KL, along with several commonly-used divergences~\citep{amari16,cichocki15}.
We can easily verify that $D_\alpha(\tilde{p}:\tilde{q})\ge0$, with $D_\alpha(\tilde{p}:\tilde{q})=0$
if and only if $\tilde{p}=\tilde{q}$.
As a generalization of the KL divergence, the $\alpha$-divergence has been applied into machine learning.
Usually, $\alpha$ is a hyper-parameter of the learning machine that can be tuned.
See \eg~\cite{narayan15,li16} for recent examples.

We need the following property of $\alpha$-divergence in our developments.
\begin{lemma}[auto-normalizing]\label{thm:normalize}
Given a probability distribution $p(\bm{y})$ and a positive measure $s(\bm{y})$
so that $\int p(\bm{y}) \dy = 1$ and $0 < \int s(\bm{y}) \dy < \infty$,
the optimal $\gamma\in\Re^+\defeq(0,\infty)$
minimizing $D_{\alpha}(p\,:\,\gamma{s})$ has the form
\begin{equation*}
\gamma^\star \defeq \argmin_{\gamma\in\Re^+}
D_\alpha (p:\gamma s) =
\left(
\frac{\int p^{\alpha}(\bm{y}) s^{1-\alpha}(\bm{y}) \dy}{\int s(\bm{y}) \dy}
\right)^{1/\alpha},
\end{equation*}
and the corresponding divergence reduces to
\begin{equation*}
D_\alpha(p\,:\,\gamma^\star s)
=
\frac{1}{1-\alpha}
\left[ 1- \left(
\int p^{\alpha}(\bm{y}) q^{1-\alpha}(\bm{y}) \dy
\right)^{1/\alpha} \right],
\end{equation*}
where $q(\bm{y})\defeq s(\bm{y})/\int s(\bm{y})\dy$ is the normalized density
\wrt the positive measure $s(\bm{y})$.
\end{lemma}
The proofs of our formal statements are in the chapter appendix.
Note the expression of $D_\alpha(p\,:\,\gamma^\star s)$ is \emph{not} the same as $D_\alpha(p\,:\,q)$
in \cref{eq:alphadivnormalized}.
For a given $\alpha$, it is clear that they both
reduce to compute the Hellinger integral $\int p^{\alpha}(\bm{y}) q^{1-\alpha}(\bm{y}) \dy$.
Because $t^{1/\alpha}$ is a monotonic transformation of $t/\alpha$ for $t\in\Re^+$,
$D_\alpha(p\,:\,\gamma^\star s)$ is a deterministic monotonic function of $D_\alpha(p\,:\,q)$.

Intuitively, $\gamma$ plays the role of an ``auto-normalizer''.
If $\bm{y}=i$ is a discrete random variable,
then minimizing $D_{\alpha}(p:\gamma{}s)$ boils down to
minimizing the $\alpha$-divergence between
two discrete probability distributions:
$p_i$ and $q_i=s_i/\sum_{i}s_i$.

In the statement of \cref{thm:normalize}, we ignore the special cases when $\alpha=0$ or
$\alpha=1$ for simpler expressions.
One can retrieve these special cases by applying L'H\^{o}spital's rule, which is straightforward.
Unless the cases for $\alpha=0$ or $\alpha=1$ are explicitly discussed,
similar treatment of the parameter $\alpha$ is applied throughout the paper,
\begin{remark}
If $\alpha=1$, then $\lambda^\star=1/\int s(\bm{y})\dy$ and $\lambda^\star s(\bm{y}) = q(\bm{y})$.
In this special case,
\begin{equation*}
    D_1(p\,:\,\gamma^\star s) = D_1(p\,:\, q) = \kl(p\,:\,q).
\end{equation*}
In general, $\forall\alpha\in\Re$,
as $\lambda^\star$ is the minimizer of $D_\alpha(p\,:\,\gamma{} s)$, we have
\begin{equation*}
    0 \le D_\alpha(p\,:\,\gamma^\star s) \le D_\alpha(p\,:\, q).
\end{equation*}
\end{remark}

\begin{remark}
For a given $\alpha$, minimizing $D_{\alpha}(p\,:\,q_{\bm\theta})$, where
$q_{\bm\theta}$ is a parametric family of distributions with parameters $\bm\theta$,
is equivalent to minimizing $D_{\alpha}(p\,:\,\gamma{s}_{\bm\theta})$
\wrt both $\gamma$ and $\bm\theta$:
\begin{equation*}
\bm\theta^\star
    \defeq
\argmin_{\bm\theta} D_{\alpha}(p\,:\,q_{\bm\theta})
    =
\argmin_{\bm\theta} \min_{\gamma\in\Re^+} D_{\alpha}(p\,:\,\gamma{s}_{\bm\theta}),
\end{equation*}
where $s_{\bm\theta}$ is the unnormalized positive measure \wrt $q_{\bm\theta}$.
\end{remark}
In machine learning, $D_{\alpha}(p\,:\,q_{\bm\theta})$ is often used
as the \emph{loss} to be minimized \wrt the free parameters $\bm\theta$.
$D_{\alpha}(p\,:\,\gamma{s}_{\bm\theta})$ is a surrogate function of
$D_{\alpha}(p\,:\,q_{\bm\theta})$ after some deterministic transformations
depending on $\alpha$. In parametric learning, it may be favorable to
minimize $D_{\alpha}(p\,:\,\gamma{s}_{\bm\theta})$ for its simpler expression
without the terms related to the normalizer $\int s_{\bm\theta}(\bm{y})\dy$.
One may discover the Hessian of $D_{\alpha}(p\,:\,q_{\bm\theta})$ \wrt
$\bm\theta$ and the Hessian of $D_{\alpha}(p\,:\,\gamma{s}_{\bm\theta})$ \wrt
$\bm\theta$ are different. Therefore these two different loss landscapes
are different. These application scenarios are beyond the scope of the current paper.

\section{$\alpha$-Discrepancy of an embedding}\label{sec:embed}

How to measure the amount of \emph{imposed information}, or \emph{complexity},
of a mapping $f:\,\calY\to\calX$
between two manifolds $\calY$ and $\calX$, in such a way that is independent to the observations?
Taking dimensionality reduction as an example, the central problem
is to find such a ``good'' mapping $f$, which not only fits well
the observations but also is somehow simple in the sense that it is less curved.
In this case, $f$ is from a latent space $\calY^{d}$ that is usually a low dimensional
Euclidean space\footnote{In this paper, a $d$-dimensional real manifold is denoted by $\mathcal{M}^d$,
or simply $\calM$ where the superscript (dimensionality) is omitted.}, to an
observable space  $\calX$, which often has a high
dimensionality. For example, the index mapping ${i}\to\bm{x}_i$ is usually considered as
``not good'' for dimensionality reduction, because only the information
regarding the order of the samples (if such information exists) is preserved in the embedding,
and all other observed information is lost through the highly curved $f$.

\subsection*{$\alpha$-Discrepancy}

We define such an intrinsic loss of information and discuss its properties.
For the convenience of analysis, we make the following assumptions:
\begin{description}
\item[Assumption \ding{182}] $\calX$ is equipped with a Riemannian metric $M(\bm{x})$,
    \ie a covariant symmetric tensor that is positive definite and varies soothingly \wrt $\bm{x}\in\calX$.

\item[Assumption \ding{183}] The latent space $\calY\defeq\Re^d$ is a $d$-dimensional
    Euclidean space endowed with a positive similarity $s_{\bm{y}}(\bar{\bm{y}})$: $\calY^2\to\Re^{+}$,
    where $\bm{y}\in\calY$ is called a \emph{reference point},
    and $\bar{\bm{y}}\in\calY$ is called a \emph{neighbour}.
    This measure has simple closed form, and satisfies
    \begin{equation*}
    \forall{\bm{y}\in\calY},\quad\quad{} 0 < \int s_{\bm{y}}(\bar{\bm{y}}) \dbary < \infty.
    \end{equation*}

\item[Assumption \ding{184}] There exists a differentiable mapping $f:\,\calY\to\calX$,
    whose Jacobian $J(\bm{y})$, or simply denoted as $J$, can be computed. Depending on our analysis, we may
    further assume the Jacobian $J$ has full column rank everywhere on
    $\calY$\footnote{Such an embedding is known as an ``immersion''~\citep{jost11}.}.
    Such an assumption excludes neural network
    architectures which reduce $\calY$'s dimensionality in some hidden layer
    in between $\calY$ and $\calX$.

\item[Assumption \ding{185}] The following generation process:
A latent point $\bm{y}$ is drawn from a prior distribution $\calU(\bm{y})$ defined on $\calY$.
Its corresponding observed point is $\bm{x} = f(\bm{y})\in\calX$.
This $\calU(\bm{y})$ can be Gaussian, uniform, \etc.
\end{description}

The mapping $f: \calY\to\calX$ induces a metric tensor of $\calY$,
given by the pullback metric~\citep{jost11} $M(\bm{y}) \defeq J^\T(\bm{y}) M(f(\bm{y})) J(\bm{y})$,
where $J(\bm{y})\defeq\frac{\partial\bm{x}}{\partial\bm{y}}\vert$ is the Jacobian matrix at $\bm{y}$.
We abuse $M$ to denote both the Riemannian metric on $\calX$ and the pullback metric on $\calY$.
Informally, it means that the geometry of $\calY$ is
based on how the image $f(\calY)$ is curved inside the ambient space $\calX$.
By definition, $M(\bm{y})$ is only positive semi-definite and may have zero
eigenvalues. Moreover, $M(\bm{y})$ may not be smooth when $\bm{y}$ varies in $\calY$.
Therefore, further assumptions are needed to make the pullback metric to be a
valid Riemannian metric. The notion of pullback metric is applied in machine
learning~\citep{lebanon02,sun14,tosi14} including deep
learning~\citep{arvanitidis18,hauberg,fishergcn,sun20}. Related work will be discussed
later in \cref{sec:connections}.

In order to apply information geometric measurements, we consider a probability
density defined \wrt this induced geometry, given by
\begin{equation*}
p_{\bm{y}}(\bar{\bm{y}})
\defeq
G\left( \bar{\bm{y}}\,\vert\,\bm{y}, J^\T(\bm{y}) M(f(\bm{y})) J(\bm{y}) \right)
\propto
\exp\left(
-\frac{1}{2}(\bar{\bm{y}}-\bm{y})^\T J^\T(\bm{y}) M(f(\bm{y})) J(\bm{y}) (\bar{\bm{y}}-\bm{y})
\right),
\end{equation*}
where $G(\cdot\,\vert\,\bm\mu,\,\Sigma)$ denotes the
multivariate Gaussian distribution centered at $\bm\mu$
with precision matrix $\Sigma$.
Formally, $p_{\bm{y}}(\bar{\bm{y}})$ is a probability distribution locally
defined on the tangent space $\calT_{\bm{y}}\calY$.
If the pullback geometry is Riemannian, a neighbourhood of $\bm{y}$ is given by
$\{\exp_{\bm{y}}(\bar{\bm{y}})\,:\,\bar{\bm{y}}\sim p_{\bm{y}}(\bar{\bm{y}})\}$,
where $\exp_{\bm{y}}:\,\calT_{\bm{y}}\calY\to\calY$ is the Riemannian exponential map.
In this paper, we relax the formal requirements and perceive
$p_{\bm{y}}(\bar{\bm{y}})$ as a local neighbourhood distribution around
$\bm{y}\in\calY$.
Here, we define similarities based on a given Riemannian metric,
which is the reverse treatment of inducing metrics based on kernels (see 11.2.4 of~\cite{amari16}).
Gaussian distributions on Riemannian manifolds are formally defined~\citep{pennec,said}.

On the other hand, by assumption \ding{183} the latent space $\calY$ is endowed with a
positive similarity measure $s_{\bm{y}}(\bar{\bm{y}})$, where
$\bar{\bm{y}},\bm{y}\in\calY$. Typical choices of $s_{\bm{y}}(\bar{\bm{y}})$ can be
\begin{equation*}
s_{\bm{y}}(\bar{\bm{y}})
=
\exp\left( -\frac{1}{2}\Vert \bar{\bm{y}}-\bm{y}\Vert^2 \right)
\text{~~~~or~~~~}
s_{\bm{y}}(\bar{\bm{y}})
=
\left(1+\frac{\Vert \bar{\bm{y}}-\bm{y}\Vert^2}{\kappa}\right)^{-\frac{\kappa+1}{2}}
(\kappa\ge1),
\end{equation*}
where $\Vert\cdot\Vert$ is the Euclidean norm.
Both of those choices are isotropic and decrease as the distance $\Vert
\bar{\bm{y}}-\bm{y}\Vert$ increases. The meaning of $s_{\bm{y}}(\bar{\bm{y}})$
is a local neighbourhood of $\bm{y}\in\calY$ based on the intrinsic geometry of
$\calY$.

We try to align these two geometries by comparing the local probability distribution
$p_{\bm{y}}(\bar{\bm{y}})$ and the positive measure $s_{\bm{y}}(\bar{\bm{y}})$,
which can be gauged by the $\alpha$-divergence (a ``distance'' between positive measures)
introduced in \cref{sec:alpha}.
The underlying assumption is that, under a simple mapping $f$, these two
different neighbourhoods should be similar.
On the other hand, after a complex mapping $f$, the neighbourhood structure is
likely to be destroyed. The complexity of $f:\,\calY\to\calX$ can be defined as follows.
\begin{mydefinition}[Embedding $\alpha$-discrepancy]\label{def:alpha}
With respect to the assumptions \ding{182}, \ding{183}, \ding{184}
the local $\alpha$-discrepancy of $f:\,\calY\to\calX$ is a function on $\calY$:
\begin{equation*}
\calD_{\alpha,f}(\bm{y})
\defeq
\inf_{\gamma\in{}C(\calY)}
D_\alpha( p_{\bm{y}}:\gamma(\bm{y}) s_{\bm{y}} ),
\end{equation*}
where
$C(\calY)\defeq\{\gamma\,:\,\calY\to\Re^+\,:\,\forall{\bm{y}}\in\calY, \gamma(\bm{y})>0\}$
is the set of positive functions on $\calY$.
If further assumption \ding{185} if true, the (global) $\alpha$-discrepancy
of $f$ is
\begin{equation*}
\calD_{\alpha,f} \defeq E_{\calU(\bm{y})} \calD_{\alpha,f}(\bm{y}),
\end{equation*}
where $E_{\calU(\bm{y})}(\cdot)$ denotes the expectation \wrt $\calU(\bm{y})$.
\end{mydefinition}
\begin{remark}
    An alternative definition of the $\alpha$-discrepancy can be
    based on the $\alpha$-divergence between two normalized densities
    $p_{\bm{y}}$ and $q_{\bm{y}}$.
    In this paper, we choose the more general \cref{def:alpha},
    as $\gamma(\bm{y})$ can be automatically normalized based on
    \cref{thm:normalize}.
\end{remark}
As $\calD_{\alpha,f}$ is measured against the mapping $f:\,\calY\to\calX$
instead of two probability measures,
we use the term ``discrepancy'' (denoted as $\calD$) instead of ``divergence''
(denoted as $D$).
The $\alpha$-discrepancy $\calD_{\alpha,f}(\bm{y})$ is a scalar function on the
manifold $\calY$, as its value varies \wrt the reference point $\bm{y}\in\calY$.
The global discrepancy
$\calD_{\alpha,f}$ is simply a weighted average of the local
discrepancy \wrt some prior distribution $\calU(\bm{y})$.
Obviously, from the non-negativity of $\alpha$-divergence,
$\forall\bm{y}\in\calY$, $\forall\alpha\in\Re$, and for any differentiable $f$,
we have $\calD_{\alpha,f}(\bm{y})\ge0$.

As the $\alpha$-divergence is in general unbounded, $\calD_{\alpha,f}(\bm{y})$ is also unbounded.
Note, when $\alpha=1/2$ the $\alpha$-divergence becomes the Hellinger distance,
which is a \emph{bounded} metric. Depending on the application, it maybe favored
as it allows $p_{\bm{y}}(\bar{\bm{y}})$ and $q_{\bm{y}}(\bar{\bm{y}})$
to be zero, and therefore has better numerical stability.

To interpret the meaning of the proposed discrepancy value, we have the following basic property.
\begin{proposition}\label{thm:isometry}
Assume that
$s_{\bm{y}}( \bar{\bm{y}} ) \defeq \exp\left(-\frac{1}{2}\Vert\bar{\bm{y}}-\bm{y}\Vert^2\right)$.
$\forall\alpha\in\Re$,
any isometry $f:\,\calY\to\calX$ (if exists) is an optimal solution of $\min_{f} \calD_{\alpha,f}$.
\end{proposition}

Notice that an ``isometric embedding'' is defined by such mappings where the induced metric $M(\bm{y})$
is everywhere equivalent to the intrinsic metric of the Euclidean space $\calY$
given by the identity matrix $I$.
To gain some intuitions of $\calD_{\alpha,f}$,
consider the special case, where $\dim(\calX)=\dim(\calY)$ and
$f$ is a change of coordinates. Minimizing $\calD_{\alpha,f}$ \wrt $f$ helps to find
a ``good'' coordinate system where the metric tensor is best aligned to $I$.
Consider $M(\bm{x})$ is non-isometric and is small along the
directions of the data point cloud~\citep{lebanon02},
so that geodesics walk along the observed data points.
Minimizing the $\alpha$-discrepancy means to transform the
coordinate system so that unit balls centered around the data points
are like ``pancakes'' along the data manifold.

The $\alpha$-divergence belongs to the broader $f$-divergence~\citep{csiszar67} family
and therefore inherits the primitives. By the invariance of $f$-divergence~\citep{amari16},
its value does not change \wrt coordinate transformation of the support.
We have the following property of \cref{def:alpha}.
\begin{proposition}\label{thm:invariant}
$\calD_{\alpha,f}(\bm{y})$ (and therefore $\calD_{\alpha,f}$)
is invariant \wrt any reparameterization of the observation space $\calX$ or the latent space $\calY$.
In other words, for any given diffeomorphisms $\Phi_\calX:\,\calX\to\calX$ and $\Phi_\calY:\calY\to\calY$,
we have
\begin{equation*}
\forall\bm{y}\in\calY,
\quad
\calD_{\alpha, f}(\bm{y})
    =
\calD_{\alpha, \Phi_\calX\circ{}f\circ{}\Phi_\calY}(\Phi^{-1}_\calY(\bm{y})).
\end{equation*}
\end{proposition}

Indeed, consider the observation space is reparameterized to a new coordinate system $\bm{x}'$,
where the Jacobian of $\bm{x}\to\bm{x}'$ is given by $J_x\defeq{J}_x(\bm{x})$. Then
the pullback metric becomes
$M(\bm{y}) = J^\T J_x^\T M(\bm{x}') J_x J=J^\T M(\bm{x}) J$.
Therefore, the $\alpha$-discrepancy is an intrinsic measure
solely determined by $\alpha$, $f$ and the geometry of $\calX$ and $\calY$,
and is regardless of the choice of the coordinate system.

In order to examine the analytic expression of the $\alpha$-discrepancy,
we have the following theorem.
\begin{theorem}\label{thm:jmj}
If $s_{\bm{y}}(\bar{\bm{y}})
\defeq
\exp\left( -\frac{1}{2}\Vert\bar{\bm{y}} - \bm{y}\Vert^2 \right)$,
then
\begin{align}
\calD_{\alpha,f}(\bm{y})
=
&
\frac{1}{1-\alpha}\bigg[
1-
\frac{\vert{J}^\T(\bm{y}) {M}(f(\bm{y})) J(\bm{y}) \vert^{1/2}}
     { \left\vert\alpha
       J^\T(\bm{y}) {M}(f(\bm{y})) J(\bm{y})
       + (1-\alpha) {I}\right\vert^{1/2\alpha}
     } \bigg],\nonumber
\end{align}
where $I$ is the identity matrix.
\end{theorem}
Observe that, for general values of $\alpha$, the $\alpha$-discrepancy can be
well-defined even for singular $J(\bm{y})$.

\begin{remark}
We give explicitly the expressions of $\calD_{\alpha,f}(\bm{y})$ for $\alpha=0\text{~or~}1$:
\begin{align*}
\calD_{0,f}(\bm{y})
& =
1-\exp\left[
 \frac{1}{2} \log\left\vert J^\T(\bm{y}) M(f(\bm{y})) J(\bm{y}) \right\vert
 -\frac{1}{2} \tr\left( J^\T(\bm{y}) M(f(\bm{y})) J(\bm{y}) \right)
 +\frac{d}{2}
 \right],\\
\calD_{1,f}(\bm{y})
 &=
 \frac{1}{2}\log\vert J^\T(\bm{y}) M(f(\bm{y})) J(\bm{y}) \vert
+\frac{1}{2} \tr\left( (J^\T(\bm{y}) M(f(\bm{y})) J(\bm{y}) )^{-1} \right)
-\frac{d}{2},
\end{align*}
where $\tr(\cdot)$ denotes the trace, and $d$ is the dimensionality of $\calY$.
\end{remark}

\begin{remark}\label{remark:logdet}
If $s_{\bm{y}}(\bar{\bm{y}})
\defeq
\exp\left( -\frac{1}{2}\Vert\bar{\bm{y}} - \bm{y}\Vert^2 \right)$,
it is easy to see that $\calD_{\alpha,f}(\bm{y})$ is a dissimilarity measure between
$M(\bm{y})$ and $I$,
while $\calD_{0,f}(\bm{y})$ and $\calD_{1,f}(\bm{y})$
is in the form of a LogDet divergence~\citep{cichocki15}
between two positive definite matrices $A$ and $B$
(up to a monotonic transformation):
\begin{equation*}
\mathrm{LogDet}(A,B) = \tr(A B^{-1}) - \log\vert AB^{-1} \vert.
\end{equation*}
\end{remark}

The following corollary is a ``spectrum version'' of \cref{thm:jmj}.
\begin{corollary}\label{thm:spectrum}
If $\forall{\bm{y}}$, $M(f(\bm{y}))=I$, and $s_{\bm{y}}(\bar{\bm{y}})
\defeq
\exp\left( -\frac{1}{2}\Vert\bar{\bm{y}} - \bm{y}\Vert^2 \right)$, then
\begin{equation}
\calD_{\alpha,f}(\bm{y})
=
\frac{1}{1-\alpha}
\left[
    1- \prod_{i=1}^d \frac{\tau_i(\bm{y})}{ (1+\alpha\tau^2_i(\bm{y})-\alpha)^{1/2\alpha} }
\right],
\end{equation}
where $\tau_i(\bm{y})$ ($i=1,\cdots,d$) denotes the singular values of $J(\bm{y})$.
\end{corollary}

It is well known (see \eg \cite{jose16,cichocki15}) that the $\alpha$-divergence
presents different properties \wrt different settings of $\alpha$.
For example, when $\alpha\ge1$, minimizing $D_\alpha(p:q)$
will enforce $q>0$ (zero-avoiding) whenever $p>0$.
Similarly, we have the following remark on the setting of $\alpha$.
\begin{remark}
  Based on \cref{remark:logdet} and \cref{thm:spectrum},
  the value of $\calD_{\alpha,f}(\bm{y})$ could be sensitive to the settings of $\alpha$.
\begin{itemize}
 \item
  If $\alpha\to1$, small singular values of $J(\bm{y})$ that are close to
  0 lead to high discrepancy. Minimizing $\calD_{\alpha,f}(\bm{y})$ helps avoid
  singularity.
\item
  If $\alpha\to0$, large singular values of $J(\bm{y})$ that are larger than 1
  lead to high discrepancy.
  Minimizing $\calD_{\alpha,f}(\bm{y})$ favors contractive mappings~\citep{rifai11}
  where the scale of the Jacobian is constrained.
\end{itemize}
\end{remark}

\section{Empirical $\alpha$-discrepancy}\label{sec:empirical}

In this section, we show that the proposed $\alpha$-discrepancy is computationally friendly.
By definition, the global $\alpha$-discrepancy $\calD_{\alpha,f}$
is an expectation of the local $\alpha$-discrepancy $\calD_{\alpha,f}(\bm{y})$
\wrt a given prior distribution $\calU(\bm{y})$.
Using $m$ \iid samples $\{\bm{y}_i\}_{i=1}^m\sim\calU(\bm{y})$, we can estimate
\begin{equation*}
\calD_{\alpha,f}
\approx
\frac{1}{m} \sum_{i=1}^m \calD_{\alpha,f}(\bm{y}_i).
\end{equation*}
In machine learning problems, usually we are already given a set of observed points
$\{\bm{x}_i\}_{i=1}^m\subset\calX$, whose latent representations $\{\bm{y}_i\}_{i=1}^m$
can be ``learned''. We may choose $\calU(\bm{y})$ to be the empirical distribution
so that $\calU(\bm{y}) \defeq \frac{1}{m}\sum_{i=1}^m \delta(\bm{y}-\bm{y}_i)$,
where $\delta(\cdot)$ is the Dirac delta function.
Therefore the problem reduces to compute $\calD_{\alpha,f}(\bm{y})$
based on some given $\alpha$, $f$ and $\bm{y}$.

By definition, $\calD_{\alpha,f}(\bm{y})$ is the $\alpha$-divergence
between a Gaussian distribution $p_{\bm{y}}$ and a positive measure
$\lambda s_{\bm{y}}$.
As $s_{\bm{y}}$ is in simple closed form by our assumption \ding{183},
and $\lambda$ can be fixed based on \cref{thm:normalize},
the remaining problem is to compute the covariance matrix of $p_{\bm{y}}$
given by $J^\T(\bm{y}) M(f(\bm{y})) J(\bm{y})$,
which depends on the Jacobian $J(\bm{y})$ of the mapping $f$.
Afterwards, $\calD_{\alpha,f}(\bm{y})$ can be computed based on either \cref{thm:jmj},
or a numerical integrator.

For deep neural networks, the mapping $f$ is usually specified by a composition of
linear mappings and non-linear activation layers, so that
\begin{equation*}
f \defeq f_L\circ{}\cdots\circ{f}_2\circ{}f_1,
\end{equation*}
where $L\ge1$ is the number of layers,
$f_l(\bm{h}_l) \defeq \nu(W_l\bm{h}_l+\bm{b}_l)$, $l=1,\cdots,L$,
$\bm{h}_l$ is the vector input of the $l$'th layer, $\bm{h}_1\defeq\bm{y}$,
$W_l$ is the weight matrix, $\bm{b}_l$ is the bias terms,
and $\nu$ is an elementwise non-linear activation function,
\eg ReLU~\citep{nair10}. By the chain rule,
\begin{equation}
    J(\bm{y}) = \prod_{l=1}^L A_l(\bm{h}_l) W_l,
\end{equation}
where $A_l(\bm{h}_l)$ is a diagonal matrix with diagonal entries $\nu'(W_l\bm{h}_l+\bm{b}_l)$.
Therefore, $J(\bm{y})$ is expressed as a series of matrix multiplications.
For general neural network mapping $f$ not limited to the above
simple case, one can use the ``Jacobian'' programming interface that is highly
optimized in modern auto-differentiation frameworks~\citep{torch}.

In the following, we show that $\calD_{\alpha,f}(\bm{y})$ can be approximated by
Monte Carlo sampling.
By the definition of $\calD_{\alpha,f}(\bm{y})$,
we only need to estimate $D_{\alpha}( p_{\bm{y}}\,:\,\gamma s_{\bm{y}})$.
For a given $\bm{y}$, we draw a set of $n$ neighbours
$\{\bar{\bm{y}}_j\}_{j=1}^n \sim{R}_{\bm{y}} (\bar{\bm{y}})$,
where $R_{\bm{y}}$ is a reference probability distribution
defined on the same support as $p_{\bm{y}}$ and $s_{\bm{y}}$. Therefore
\begin{align}\label{eq:simplex}
\hat{D}_{\alpha}( p_{\bm{y}}\,:\,\gamma s_{\bm{y}})
&=
\frac{1}{1-\alpha}
+
\frac{1}{n}\sum_{j=1}^n
\left[
\frac{\gamma s_{\bm{y}}(\bar{\bm{y}}_j) }{\alpha R_{\bm{y}}(\bar{\bm{y}}_j)}
- \frac{ p^\alpha_{\bm{y}}(\bar{\bm{y}}_j) \gamma^{1-\alpha} s^{1-\alpha}_{\bm{y}}(\bar{\bm{y}}_j) }
{\alpha(1-\alpha)R_{\bm{y}}( \bar{\bm{y}}_j )}
\right]
\end{align}
gives an unbiased estimate of $D_{\alpha}(p_{\bm{y}}\,:\,\gamma s_{\bm{y}})$,
where the notation $\hat{D}_{\alpha}( p_{\bm{y}}\,:\,\gamma s_{\bm{y}})$
is abused as it depends on the samples $\{\bar{\bm{y}}_j\}_{j=1}^n$. Indeed,
\begin{align*}
E_{R_{\bm{y}}(\bar{\bm{y}})} \hat{D}_{\alpha}( p_{\bm{y}}\,:\,\gamma s_{\bm{y}})
&=
\frac{1}{1-\alpha}
+
\int R_{\bm{y}}(\bar{\bm{y}})
\left[
\frac{\gamma s_{\bm{y}}(\bar{\bm{y}}) }{\alpha R_{\bm{y}}(\bar{\bm{y}})}
- \frac{ p^\alpha_{\bm{y}}(\bar{\bm{y}}) \gamma^{1-\alpha} s^{1-\alpha}_{\bm{y}}(\bar{\bm{y}}) }
{\alpha(1-\alpha) R_{\bm{y}}(\bar{\bm{y}})}
\right]
\dbary\nonumber\\
&=
\frac{1}{1-\alpha}
+
\frac{1}{\alpha}
\int \gamma s_{\bm{y}}(\bar{\bm{y}}) \dbary
-\frac{1}{\alpha(1-\alpha)}
\int
p^\alpha_{\bm{y}}(\bar{\bm{y}}) \gamma^{1-\alpha} s^{1-\alpha}_{\bm{y}}(\bar{\bm{y}})
\dbary,
\end{align*}
which gives exactly $D_{\alpha}(p_{\bm{y}}\,:\,\gamma s_{\bm{y}})$.
By the large number law, $\hat{D}_{\alpha}( p_{\bm{y}}\,:\,\gamma s_{\bm{y}})$
converges to the $\alpha$-divergence
$D_{\alpha}( p_{\bm{y}}\,:\,\gamma s_{\bm{y}})$ as $n\to\infty$.
This holds regardless of the choice of the reference distribution $R_{\bm{y}}(\bar{\bm{y}})$.
However different $R_{\bm{y}}(\bar{\bm{y}})$ may result in different estimation variance.
The RHS (right-hand-side) of \cref{eq:simplex} is a convex function of $\gamma$,
and the optimal $\gamma^\star$ which minimizes
$\hat{D}_{\alpha}( p_{\bm{y}}\,:\,\gamma s_{\bm{y}})$
is in closed form (similar to \cref{thm:normalize}).
Then, one can estimate
\begin{equation*}
\calD_{\alpha,f}(\bm{y})
\approx
\min_{\gamma\in\Re^+} \hat{D}_{\alpha}( p_{\bm{y}}\,:\,\gamma s_{\bm{y}}).
\end{equation*}
This empirical $\alpha$-discrepancy is useful when ${D}_{\alpha}(
p_{\bm{y}}\,:\,\gamma s_{\bm{y}})$ is not available in closed form.

\section{Connections to existing methods}\label{sec:connections}

In this section, we show that machine learning based on the proposed
$\alpha$-discrepancy encompasses some well-studied techniques, and thus we
uncover a connection in between those existing methods.

\subsection*{Neighbourhood embeddings}

The neighbourhood embeddings~\citep{hinton03,cook07,maaten08,cp10,venna07,lee13,yang14}
are based on our third paradigm introduced
in~\cref{sec:intro}. Based on similar principles,
this meta family of dimensionality reduction methods
aim to find the embedding (inverse of $f\,:\calY\to\calX$) based on a given set
$\{\bm{x}_i\}\subset\calX$. In this section, we provide a unified view of
these embedding techniques based on the proposed $\alpha$-discrepancy.

Often, neighbourhood embedding are non-parametric methods, meaning that the
parametric form of the mapping $f$ (or its inverse $g$) is not explicitly
learned. Instead, the embedding points $\{\bm{y}_i=g(\bm{x}_i)\}$, which
implicitly gives $g(\cdot)$, are obtained by minimizing a loss function. They
directly generalize to parametric approaches~\citep{cp15}. Our analysis can be
applied to both cases. However, the Jacobian may be difficult to obtain for the
non-parametric mapping and requires approximation techniques.

In the empirical $\alpha$-discrepancy, we set the reference distribution
$R_{\bm{y}}$ to be uniform across the latent $\{\bm{y}_i\}$ associated
with the observed samples $\{\bm{x}_i\}$.
This is only a technical treatment which allows us
to arrive at a simple loss function
and to connect with related embedding methods.
Other choices of $R_{\bm{y}}$ lead to more general embedding approaches.
Thus, we simply set $\forall{j}$, $R_{\bm{y}}(\bar{\bm{y}}_j)=\rho>0$.

From \cref{eq:simplex}, we get
\begin{align}\label{eq:presne}
\hat{D}_\alpha ( p_{\bm{y}} : \gamma{s}_{\bm{y}} )
=
\frac{1}{1-\alpha}
+
\frac{1}{n} \sum_{j=1}^n
\left[
  \frac{ \gamma s_{\bm{y}}(\bar{\bm{y}}_j) }{\alpha \rho}
- \frac{ p_{\bm{y}}^{\alpha}( \bar{\bm{y}}_j ) \gamma^{1-\alpha} s^{1-\alpha}_{\bm{y}}(\bar{\bm{y}}_j) }
       { \alpha(1-\alpha)\rho}
\right].
\end{align}
By differentiating the RHS \wrt $\gamma$, we obtain the optimal $\gamma$:
\begin{equation*}
(\gamma^\star)^\alpha =
\frac{\sum_{j=1}^n p_{\bm{y}}^\alpha(\bar{\bm{y}}_j) s_{\bm{y}}^{1-\alpha}(\bar{\bm{y}}_j)}
{\sum_{j=1}^n s_{\bm{y}}(\bar{\bm{y}}_j)}.
\end{equation*}
Plugging $\gamma^\star$ into \cref{eq:presne}, we get
\begin{equation}\label{eq:sne}
\hat{D}_\alpha ( p_{\bm{y}} : \gamma^\star {s}_{\bm{y}} )
=
\frac{1}{1-\alpha}
-\frac{1}{(1-\alpha)\rho}
\left[
\sum_{j=1}^n p_{\bm{y}}^{\alpha}( \bar{\bm{y}}_j ) q^{1-\alpha}_{\bm{y}}(\bar{\bm{y}}_j)
\right]^{1/\alpha},
\end{equation}
where $q_{\bm{y}}(\bar{\bm{y}}_j)
= s_{\bm{y}}(\bar{\bm{y}}_j)/\sum_{j=1}^n s_{\bm{y}}(\bar{\bm{y}}_j)$.
From \cref{thm:normalize} and related discussions, the RHS of \cref{eq:sne}
is the $\alpha$-divergence between the two probability mass functions:
$p_{\bm{y}} (\bar{\bm{y}}_j )$ and $q_{\bm{y}}(\bar{\bm{y}}_j)$ up
to a monotonic transformation. Let $\alpha\to1$, then
it reduces to the KL divergence, which, depending on the choice of $s_{\bm{y}}$,
is exactly the loss function of
stochastic neighbour embedding (SNE)~\citep{hinton03} or t-SNE~\citep{maaten08}.
They are implemented by minimizing the empirical average of
$\hat{D}_\alpha( p_{\bm{y}_i} : \gamma^\star {s}_{\bm{y}_i} )$
over $\{\bm{y}_i\}$. Note $\bar{\bm{y}}_j$ are sampled according to
$R_{\bm{y}_i}$, which is uniform on the population $\{\bm{y}_i\}$. In other
words, we use each $\bm{y}_i$ as the reference point, and use all the other
samples as the neighbours.

A further algorithmic detail in SNE~\citep{hinton03} is for computing the
probabilities $p_{\bm{y}}(\bar{\bm{y}}_j)$. Essentially, SNE sets $M(\bm{x}) = \lambda(\bm{x}) I$,
where $\lambda\defeq\lambda(\bm{x})>0$ is a scalar depending on
$\bm{x}$ and can be computed based on entropy constraints.
It helps model observed data with different densities at different regions on $\calX$.
Then $p_{\bm{y}}(\bar{\bm{y}}_j)$ is computed based on
\begin{equation*}
p_{\bm{y}}(\bar{\bm{y}}_j)
\propto
\exp\left(
-\frac{1}{2}( \bar{\bm{y}}_j - \bm{y} )^\T
J^\T M(\bm{x}) J
(\bar{\bm{y}}_j - \bm{y} )\right)
\approx
\exp\left( -\frac{\lambda}{2} \Vert f(\bar{\bm{y}}_j) - f(\bm{y})) \Vert^2 \right).
\end{equation*}
As $J$ is the local linear approximation of the non-linear mapping $f$,
$\Vert J (\bar{\bm{y}}_j - \bm{y} ) \Vert \approx \Vert f(\bar{\bm{y}}_j) - f(\bm{y})) \Vert$
if $\bar{\bm{y}}_j$ and $\bm{y}$ are close enough.

Let us examine the general loss on the RHS of \cref{eq:presne}.
For simplicity, let $0<\alpha<1$.
The last term in the bracket imposes an attraction strength
between $\bm{y}$ and $\bar{\bm{y}}_j$ as it tries to maximize
the similarity $s_{\bm{y}}(\bar{\bm{y}}_j)$.
The first term in the bracket forms input-independent background repulsion~\citep{cp10},
in the sense that it does not depend on $p_{\bm{y}}(\bar{\bm{y}}_j)$,
and the similarity $s_{\bm{y}}(\bar{\bm{y}}_j)$ is to be minimized.

When learning neighbourhood embeddings, one may either apply \cref{eq:presne},
where $\gamma\in\Re^+$ is a free parameter, or \cref{eq:sne}, where $\gamma$ is
fixed to $\gamma^\star$. Most methods in this family take the latter approach.
By \cref{thm:normalize}, they are actually equivalent and have
the same global optimal solution. However, they may present different properties
during numerical optimization.

If we consider $\gamma$ as a hyper-parameter that can be set to different values
(instead of fixed to $\gamma^\star$ or set free), \cref{eq:presne} becomes a
general formulation of elastic embedding (EE)~\citep{cp10}, where the attraction
terms and the repulsion terms have customized weights, and the loss does not
correspond to the KL divergence between normalized probability distributions. In
our formulations, the reference distribution $R_{\bm{y}}$ corresponds to these
weights.

For general values of $\alpha$ as a hyper-parameter,
\cref{eq:presne} gives neighbour embeddings based on the $\alpha$-divergence~\citep{venna07,lee13,yang14}.
Other choice of the information divergence extends to broader families of embeddings~\citep{narayan15}.

\subsection*{Autoencoders}

Autoencoder networks try to learn both $f_{\bm\theta}:\,\calY\to\calX$ (the decoder)
and its corresponding projection $g_{\bm\theta}:\,\calX\to\calY$ (the encoder)
to represent a given set of observations $\{\bm{x}_i\}\subset\calX$.
The following (known) statement presents an intrinsic geometry of autoencoders.
\begin{proposition}\label{thm:pullback}
Assume \ding{182}, \ding{183}, and \ding{184} are true.
In an autoencoder network $\calX\xrightarrow{g}\calY\xrightarrow{f}\calX$,
the decoder $f$ induces a metric tensor in the latent space $\calY$,
given by $M(\bm{y}) = J_f^\T(\bm{y}) M(f(\bm{y})) J_f(\bm{y})$;
the encoder $g:\,\calX\to\calY$ induces a metric tensor in the manifold $\calZ=\calX/\sim_g$
(the quotient space \wrt the equivalent relation $\sim_g$:
$\bm{x}_1\sim_g\bm{x}_2$ if and only if $g(\bm{x}_1)=g(\bm{x}_2)$)
given by $M(\bm{z})=J_g^\T(\bm{z}) M(g(\bm{z})) J_g(\bm{z})$.
\end{proposition}
The proof (omitted) is straightforward from the definition of the pullback metric.
Note the induced metric tensor is not necessarily a Riemannian metric.
For example, in ReLU networks~\citep{nair10},
$f$ is not smooth, and the metric tensor $M(\bm{y})$ may not vary smoothly along $\calY$.
If $J_f$ has zero singular values, the metric tensor may be singular and
therefore does not satisfy the conditions of a Riemannian metric.
The second part of \cref{thm:pullback} requires the quotient space
$\calZ$ to have a smooth manifold structure, which is not guaranteed.

Let us assume the input space $\calX$ is Euclidean and $M(\bm{x})=I$,
$\forall{\bm{x}}\in\calX$. The decoder induced metric is simply
$J_f^\T(\bm{y}) J_f(\bm{y})$.
Depending on the choice of $s_{\bm{y}}$, the $\alpha$-discrepancy
is a function of the matrix $J_f^\T(\bm{y}) J_f(\bm{y})$
and can have simple expressions
such as the formula of $\calD_{\alpha,f}$ in \cref{thm:jmj}.
Therefore the $\alpha$-discrepancy can be potentially used as
an information theoretical regularizer for deep autoencoders to penalize ``complex'' $f_{\bm\theta}$.
Previous works on regularizing autoencoders~\citep{rifai11}
based on the Jacobian matrix can therefore be connected with our definition.
If $s_{\bm{y}}(\bar{\bm{y}}) \defeq \exp\left( -\frac{1}{2}\Vert\bar{\bm{y}} - \bm{y}\Vert^2 \right)$,
minimizing the $\alpha$-discrepancy
pushes $J_f^{\T}(\bm{y})J_f(\bm{y})$ towards $I$.
A small value of $\alpha$-discrepancy means that the neural network
mapping $f_{\bm\theta}$ is locally an orthogonal transformation, which is
considered to have a low complexity. By our discussions in the end of \cref{sec:embed},
different settings of $\alpha$ lead to different
complexity measures based on the spectrum of $J_f^{\T}(\bm{y})J_f(\bm{y})$.
It provides flexible tools to regularize the learning of deep autoencoders.

As a popular deep generative model, variational autoencoders (VAEs)~\citep{kingma14}
learns a parametric model $p_{\bm\theta}(\bm{x},\bm{y})$
by maximizing an evidence lower bound of the log-likelihood of some observed data.
This is equivalent to minimizing the KL divergence (or in general $\alpha$-divergence as in~\cite{li16}) between
the true posterior distribution $p_{\bm\theta}(\bm{y}\,\vert\,\bm{x})$ and
a parametric variational distribution $q_{\bm\varphi}(\bm{y}\,\vert\,\bm{x})$
implemented by the encoder.
The loss to be minimized consists of the reconstruction error plus
a complexity term given by
$\kl\left(q_{\bm\varphi}(\bm{y}\,\vert\,\bm{x}):p(\bm{y})\right)$,
the KL divergence between the approximated posterior
$q_{\bm\varphi}(\bm{y}\,\vert\,\bm{x})$ and the prior $p(\bm{y})$.
Comparatively, our proposed $\alpha$-discrepancy is based on
two \emph{local} neighbourhood densities \wrt the pullback
geometry and the intrinsic geometry of $\calY$, respectively.
VAEs are generalized based on Riemannian geometry~\citep{weightedmanifold}.

We develop a unified perspective on the pullback metric and Bayes' rule.
Consider the simple case where both $\calX$ and $\calY$ are Euclidean spaces,
and $p(\bm{x}\,\vert\,\bm{y}) = G(\bm{x}\,\vert\,f(\bm{y}), \lambda I)$,
where $\lambda>0$. That means our deterministic mapping $f$ is randomized \wrt a covariance matrix
$\lambda^{-1}I$. By Bayes' rule,
\begin{equation*}
p(\bm{y}\,\vert\,\bm{x})
\propto
\calU(\bm{y}) p(\bm{x}\,\vert\,\bm{y})
\propto
\exp\left(
\log \calU(\bm{y})
- \frac{\lambda}{2} \Vert \bm{x}-f(\bm{y}) \Vert^2
\right)
\end{equation*}
Denote $g:\,\calX\to\calY$ to be an approximation of $f^{-1}$.
A Taylor expansion of $f(\bm{y})$ around $g(\bm{x})$ yields
\begin{align*}
p(\bm{y}\,\vert\,\bm{x})
\propto
\exp\left(
\log \calU(\bm{y})
- \frac{\lambda}{2}\
\Vert \bm{x} - f(g(\bm{x})) - J(g(\bm{x}))(\bm{y}-g(\bm{x})) \Vert^2
+ o( \Vert \bm{y}-g(\bm{x})\Vert )
\right),
\end{align*}
where $o(\cdot)$ is the little-o notation. If $\lambda$ is large enough, meaning the conditional distribution
$p(\bm{x}\,\vert\,\bm{y})$ tends to be deterministic,
then the second term will dominate. We arrive at the rough approximation
\begin{equation*}
p(\bm{y}\,\vert\,\bm{x})
\approx G\left( \bm{y}\,\vert\,g(\bm{x}),
\lambda J^\T(g(\bm{x}))J(g(\bm{x})) \right).
\end{equation*}
Hence, the covariance of the posterior is related to the pullback metric at $g(\bm{x})$.

The notion of pullback metric has been investigated in metric learning
\citep{lebanon02} and manifold learning~\citep{sun14}.
Latent space
geometry is studied in Gaussian process latent variable models~\citep{tosi14}.
In the realm of deep learning,
latent manifolds learned by deep generative models (including VAEs) is explored
\citep{arvanitidis18,hauberg,generative}. In this context, stochastic metric
tensor and the expected metric are investigated \citep{arvanitidis18,hauberg}.
Notice that we only consider deterministic metrics, as our mapping $f$
is deterministic. A separate neural network can be used to learn the metric tensor
\citep{kuhnel18}.
Information geometry and the pullback metric have been applied to graph neural networks~\citep{fishergcn}.
These previous efforts on deep generative models focus on study
the geometric measurements (\eg geodesics, curvature, and means) in the latent space
obtained in deep learning. In contrast, our objective is to derive information
theoretical complexity measures of a mapping $f$.

\section{Conclusion and extensions}\label{sec:con}

We study the fundamental problem on how to measure the information carried by a
non-linear mapping $f$ from a latent space $\calY$ to the observation space $\calX$.
We define the concept of $\alpha$-discrepancy, a one-parameter family of
information theoretical measurements given by a function
$\calD_{\alpha,f}(\bm{y})$ on the latent space $\calY$,
where $\alpha\in\Re$.
Intuitively, $\calD_{\alpha,f}(\bm{y})$ measures
the discrepancy between $f$ and an isometry locally at $\forall\bm{y}\in\calY$.
Our definition is invariant to reparameterization (therefore intrinsic) and is
independent to the empirical observations. Both neighbourhood embeddings and
deep representation learning are connected with this concept. It gives
theoretical insights and generalization of these methods.

Essentially, the proposed embedding $\alpha$-discrepancy measures how ``far''
$f$ is from an isometry when the $\alpha$-discrepancy achieves 0.
We can further extend its definition to measure how far $f$
is from a conformal mapping. We have the following definition.

\begin{definition}[Conformal $\alpha$-discrepancy]\label{def:conalpha}
With respect to the assumptions \ding{182}, \ding{183} and \ding{184},
the local conformal $\alpha$-discrepancy of $f:\,\calY\to\calX$ is a function on $\calY$:
\begin{equation}
\calC_{\alpha,f}(\bm{y})
\defeq
\inf_{\gamma,\zeta\in C(\calY)}
D_\alpha\left(
p_{\bm{y}} : \gamma(\bm{y}) s_{\bm{y}, \zeta}\right).
\end{equation}
where $s_{\bm{y},\zeta}(\bar{\bm{y}})
\defeq
\exp\left(-\frac{\zeta(\bm{y})}{2}\Vert\bar{\bm{y}} - \bm{y}\Vert^2\right)$.
The conformal $\alpha$-discrepancy \wrt assumption \ding{185} is
\begin{equation*}
\calC_{\alpha,f}
\defeq
E_{\calU(\bm{y})} \calC_{\alpha,f}(\bm{y}).
\end{equation*}
\end{definition}
Note $s$ is abused to denote both $s_{\bm{y}}$ in \cref{def:alpha} and $s_{\bm{y},\zeta}$ in \cref{def:conalpha}.
For a given $\bm{y}\in\calY$, the conformal $\alpha$-discrepancy is the
minimal $\alpha$-divergence
between $p_{\bm{y}}$ and $\gamma(\bm{y})
\exp\left(-\frac{\zeta(\bm{y})}{2}\Vert\bar{\bm{y}} - \bm{y}\Vert^2\right)$,
where $\gamma(\bm{y})$ and $\zeta(\bm{y})$ are free parameters depending on $\bm{y}$.
In practice, one may constrain $\zeta(\bm{y})$ to vary inside a sub-range of $\Re^+$
to avoid trivial or undesired solutions.

If $s_{\bm{y}}(\bar{\bm{y}})\defeq\exp(-\frac{1}{2}\Vert\bar{\bm{y}}-\bm{y}\Vert^2)$
is used to compute $\calD_{\alpha,f}(\bm{y})$, we have the following relationship
by definition:
\begin{equation*}
0 \le \calC_{\alpha,f}(\bm{y}) \le \calD_{\alpha,f}(\bm{y}).
\end{equation*}
Intuitively, the positive measure $\gamma(\bm{y}) s_{\bm{y},\zeta}$ is more
flexible than $\gamma(\bm{y}) s_{\bm{y}}$ and has one additional free parameter
$\zeta(\bm{y})$.
Similar to \cref{thm:isometry}, a small value of $\calC_{\alpha,f}$ indicates
that $f$ is close to a conformal mapping; and vice versa.
$\calC_{\alpha,f}$ is useful to explore learning objectives for
conformal manifold learning. It is also useful as a regularizer for deep learning,
enforcing a neural network mapping $f_{\bm\theta}$ to be conformal-like.
Related theoretical statements are similar to $\calD_{\alpha,f}$ and are omitted for brevity.

Consider the empirical $\alpha$-discrepancy in \cref{sec:empirical}. We can
choose the reference distribution
$R(\bar{\bm{y}})=q_{\bm{y}}(\bar{\bm{y}}) \defeq s_{\bm{y}}(\bar{\bm{y}})/\int s_{\bm{y}}(\bar{\bm{y}}) \dy$,
leading to another approximation of \cref{def:alpha} (as an alternative empirical $\alpha$-discrepancy):
\begin{equation}
\hat{D}_{\alpha}( p_{\bm{y}}\,:\,\gamma s_{\bm{y}})
=
\frac{1}{1-\alpha}
+
\frac{1}{n}\sum_{j=1}^n
\left[
\frac{\gamma}{\alpha}
-
\frac{ \gamma^{1-\alpha} }{\alpha(1-\alpha)}
\left(
\frac{p_{\bm{y}}(\bar{\bm{y}}_j)}{s_{\bm{y}}(\bar{\bm{y}}_j)}
\right)^{\alpha}
\right] \int s_{\bm{y}}(\bar{\bm{y}})\dy.
\end{equation}
This approximation could be favored, as the RHS has a simple expression,
and $\bar{\bm{y}}_j$ can be sampled according to
a simple latent distribution, \eg an isotropic Gaussian distribution centered
around $\bm{y}$. $\hat{D}_{\alpha}( p_{\bm{y}}\,:\,\gamma s_{\bm{y}})$ can be
evaluated by the reparameterization trick~\citep{kingma14} as
$\bar{\bm{y}}_j = \bm{y}+\bm\epsilon$, where $\bm\epsilon$ is drawn from a
zero-centered Gaussian distribution.

In certain applications, it may be reasonable to assume that $\calX$ and/or
$\calY$ is a statistical manifold (space of probability distributions).
For example, for a classifier deep neural network, the output space $\calX$
is the simplex $\Delta^d=\{\bm{x}\,:\,\sum_{i=1}^{d+1} x_i=1;~x_i>0, \forall{i}\}$.
The unique metric of statistical manifold is given by the Fisher information
metric. Thus the definition of $\alpha$-discrepancy can be extended by using
such a geometry.

\section*{Acknowledgments}

The author thank the anonymous reviewers for the insightful comments and timely reviews.

\bibliographystyle{abbrvnat}

\newpage
\appendix

\addcontentsline{toc}{section}{Appendix}
\part{Appendix}
\parttoc

In the following we provide outline proofs of the theoretical statements.

\section{Proof of \Cref{thm:normalize}}

\begin{proof}
We have
\begin{equation}\label{eq:proof1}
D_\alpha(p:\gamma s)
=
\frac{1}{\alpha(1-\alpha)}
\int \left[
\alpha p(\bm{y}) + (1-\alpha) \gamma s(\bm{y})
- p^{\alpha}(\bm{y}) \gamma^{1-\alpha} s^{1-\alpha}(\bm{y})
\right] \dy.
\end{equation}
It is easy to see that the RHS is a convex function \wrt $\gamma$.
Therefore the optimal $\gamma^\star$ can be obtained by solving
\begin{equation*}
\frac{\partial D_{\alpha}(p:\gamma s)}{\partial \gamma}
=
\frac{1}{\alpha(1-\alpha)}
\int \left[
(1-\alpha) s(\bm{y})
-(1-\alpha) p^{\alpha}(\bm{y}) \gamma^{-\alpha} s^{1-\alpha}(\bm{y})
\right] \dy = 0,
\end{equation*}
which reduces to
\begin{equation*}
\int s(\bm{y}) \dy
= (\gamma^\star)^{-\alpha} \int p^{\alpha}(\bm{y}) s^{1-\alpha}(\bm{y}) \dy.
\end{equation*}
Therefore
\begin{equation*}
(\gamma^\star)^{\alpha} =
\frac{\int p^{\alpha}(\bm{y}) s^{1-\alpha}(\bm{y}) \dy}{ \int s(\bm{y}) \dy },
\end{equation*}
and
\begin{equation*}
\gamma^\star =
\left(
\frac{\int p^{\alpha}(\bm{y}) s^{1-\alpha}(\bm{y}) \dy}{ \int s(\bm{y}) \dy }
\right)^{\frac{1}{\alpha}}.
\end{equation*}
From the above derivations, we have
\begin{equation*}
\int \gamma^\star s(\bm{y}) \dy
= (\gamma^\star)^{1-\alpha} \int p^{\alpha}(\bm{y}) s^{1-\alpha}(\bm{y}) \dy
= \int p^{\alpha}(\bm{y}) (\gamma^\star)^{1-\alpha} s^{1-\alpha}(\bm{y}) \dy.
\end{equation*}
Plugging the above equation into \cref{eq:proof1}, we get
\begin{align*}
D_\alpha(p:\gamma^\star s)
&=
\frac{1}{\alpha(1-\alpha)}
\int \left[
\alpha p(\bm{y}) + (1-\alpha) \gamma^\star s(\bm{y}) - \gamma^\star s(\bm{y})
\right] \dy\\
&=
\frac{1}{\alpha(1-\alpha)}
\int \left[ \alpha p(\bm{y}) -\alpha \gamma^\star s(\bm{y}) \right] \dy\\
&=
\frac{1}{1-\alpha} \int \left[ p(\bm{y}) - \gamma^\star s(\bm{y}) \right] \dy\\
&=
\frac{1}{1-\alpha} \int \left[ p(\bm{y}) -
\left(
\frac{\int p^{\alpha}(\bm{y}) s^{1-\alpha}(\bm{y}) \dy}{ \int s(\bm{y}) \dy }
\right)^{1/\alpha} s(\bm{y}) \right] \dy\\
&=
\frac{1}{1-\alpha}
\left[
1-
\left(
\frac{\int p^{\alpha}(\bm{y}) s^{1-\alpha}(\bm{y}) \dy}{ \int s(\bm{y}) \dy }
\right)^{1/\alpha}
\int s(\bm{y}) \dy
\right]\\
&=
\frac{1}{1-\alpha}
\left[
1-
\left(
\frac{\int p^{\alpha}(\bm{y}) s^{1-\alpha}(\bm{y}) \dy}{ (\int s(\bm{y}) \dy)^{1-\alpha} }
\right)^{1/\alpha}
\right]\\
&=
\frac{1}{1-\alpha}
\left[
1-
\left(
\int p^{\alpha}(\bm{y}) \left(\frac{s(\bm{y})}{\int s(\bm{y})}\right)^{1-\alpha} \dy
\right)^{1/\alpha}
\right]\\
&=
\frac{1}{1-\alpha}
\left[
1- \left( \int p^{\alpha}(\bm{y}) q^{1-\alpha}(\bm{y}) \dy \right)^{1/\alpha}
\right].
\end{align*}
\end{proof}

To compare $D_\alpha(p:\gamma^\star s)$ with $D_\alpha(p:q)$, we recall from
\cref{eq:alphadivnormalized} that
\begin{equation}
D_\alpha(p:q) =
\frac{1}{\alpha(1-\alpha)} \left[
1 -\int p^{\alpha}(\bm{y}) q^{1-\alpha}(\bm{y}) \dy \right].
\end{equation}

\section{Proof of \Cref{thm:isometry}}

If $s_{\bm{y}}( \bar{\bm{y}} ) =
\exp\left(-\frac{1}{2}\Vert\bar{\bm{y}}-\bm{y}\Vert^2\right)$,
then its corresponding normalized density is
$G_{\bm{y}}( \bar{\bm{y}} )
=
G(\bar{\bm{y}}\,\vert\,\bm{y},\,I)$,
which is the multivariate Gaussian distribution with mean $\bm{y}$ and covariance matrix $I$.

By \cref{thm:normalize},
\begin{align}\label{eq:minimization}
\argmin_{f} \calD_{\alpha,f}
&=
\argmin_{f}
\int \calU(\bm{y}) \calD_{\alpha,f}(\bm{y}) \dy
\nonumber\\
&=
\argmin_{f}
\int \calU(\bm{y})
\inf_{\gamma\in{}C(\calY)} D_\alpha( p_{\bm{y}}:\gamma(\bm{y}) s_{\bm{y}} )
\dy\nonumber\\
&=
\argmin_{f}
\int \calU(\bm{y}) D_\alpha( p_{\bm{y}}: G_{\bm{y}} ) \dy.
\end{align}
In the above minimization problem, notice that $f$ only appears in the term
$p_{\bm{y}}$ on the RHS.

If $f:\,\calY\to\calX$ is an isometry, then the pullback metric
is equivalent to the Riemannian metric of $\calY$ which is given
by the identity matrix $I$. Therefore, $\forall\bm{y}\in\calY$,
\begin{equation*}
    J^\T M(\bm{x}) J = I.
\end{equation*}
Therefore, $\forall\bm{y}\in\calY$,
\begin{equation*}
p_{\bm{y}}(\bar{\bm{y}})
= G(\bar{\bm{y}}\,\vert\,\bm{y},\,J^\T M(\bm{x}) J)
= G(\bar{\bm{y}}\,\vert\,\bm{y},\,I)
= G_{\bm{y}}(\bar{\bm{y}}).
\end{equation*}
Based on the basic properties of $\alpha$ divergence,
we know that $\forall\bm{y}\in\calY$, $p_{\bm{y}}=G_{\bm{y}}$ implies
$\forall\bm{y}\in\calY$, $D_\alpha( p_{\bm{y}}: G_{\bm{y}} )=0$.
At the same time,
\begin{equation*}
\int \calU(\bm{y}) D_\alpha( p_{\bm{y}}: G_{\bm{y}} ) \dy \ge 0.
\end{equation*}
Therefore, an isometric $f$ must be the optimal solution of
\cref{eq:minimization}.

\section{Proof of \Cref{thm:invariant}}

The proof needs the \emph{invariance property of the $f$-divergence}.
Given a convex function $\sigma:\Re^+\to\Re$
and a pair of normalized probability densities $p(\bm{x})$ and $q(\bm{x})$,
their $f$-divergence is in the form
\begin{equation}\label{eq:fdiv}
D_\sigma(p:q)
\defeq
\int p(\bm{x}) \sigma\left(\frac{q(\bm{x})}{p(\bm{x})}\right) \dx.
\end{equation}
Note we use $\sigma(\cdot)$ as $f(\cdot)$ is already used to denote the generative
mapping from $\calY$ to $\calX$.
The $f$-divergence satisfies the invariance: under a coordinate transformation given by
a diffeomorphism $\Phi: \bm{x}\to\bm{x}'$, the probability density becomes $p'(\bm{x}')$.
The $f$-divergence is invariant:
\begin{equation*}
    D_\sigma\left(p'(\bm{x}'):q'(\bm{x}')\right)
    =
    D_\sigma\left(p(\bm{x}):q(\bm{x})\right).
\end{equation*}
If we set $\sigma(t)=\frac{t-t^{1-\alpha}}{\alpha(1-\alpha)}$, and plugging
in \cref{eq:fdiv}, we get exactly $D_\alpha(p:q)$ in \cref{eq:alphadivnormalized}.
Therefore the $\alpha$-divergence between two probability densities belong
to the family of $f$-divergences.

\begin{proof}
We first show that
$\calD_{\alpha,\Phi_\calX\circ\alpha}(\bm{y})=\calD_{\alpha,f}(\bm{y})$
everywhere. Consider the observation space $\calX$ is reparameterized to a new
coordinate system $\bm{x}'$ based on the diffeomorphism $\Phi_{\calX}:
\bm{x}\to\bm{x}'$.
The Riemannian metric is a covariant tensor and follows the transformation rule
\begin{equation}
J^\T_\calX(\bm{x})
M( \Phi_\calX(\bm{x}) )
J_\calX(\bm{x}) =  M(\bm{x}),
\end{equation}
where $J_\calX$ denotes the Jacobian of $\Phi_\calX$.

By the chain rule, the Jacobian of the mapping
$\Phi_\calX\circ{f}:\,\bm{y}\to\bm{x}'$ is given by
$J_\calX(f(\bm{y}))\cdot{}J(\bm{y})$. Then, the pullback metric
associated with $\Phi_\calX\circ{f}$ becomes
\begin{align*}
M(\bm{y})
&=
J^\T(\bm{y})
J^\T_\calX(f(\bm{y}))
\cdot
M(\Phi_\calX(f(\bm{y})))
\cdot
J_\calX(f(\bm{y})) J(\bm{y})
\nonumber\\
&=
J^\T(\bm{y})
\cdot
\left[
J^\T_\calX(f(\bm{y})) M(\Phi_\calX(f(\bm{y}))) J_\calX(f(\bm{y}))
\right]
\cdot
J(\bm{y})
\nonumber\\
&=
J(\bm{y})^\T M(f(\bm{y})) J(\bm{y}).
\end{align*}
In \cref{def:alpha}, $p_{\bm{y}}$ is invariant \wrt $\Phi_\calX$.
Therefore $\forall\bm{y}\in\calY$,
$\calD_{\alpha,f}(\bm{y}) = \calD_{\alpha,\Phi_\calX\circ{f}}(\bm{y})$.

In the following, we show $\forall{\bm{y}}\in\calY$,
$\calD_{\alpha,f}(\bm{y}) = \calD_{\alpha,f\circ{}\Phi_\calY}(\bm{y})$,
where $\Phi_\calY:\,\calY\to\calY$ is a diffeomorphism. By \cref{thm:normalize},
\begin{align}
\calD_{\alpha,f}(\bm{y})
=
\inf_{\gamma\in{}C(\calY)} D_\alpha( p_{\bm{y}}:\gamma(\bm{y}) s_{\bm{y}} )
\end{align}
is a monotonic function of $D_\alpha( p_{\bm{y}}\,:\,q_{\bm{y}} )$,
where $q_{\bm{y}}$ is the normalized density associated with $s_{\bm{y}}$.
By the invariance of $\alpha$-divergence, under the diffeomorphism
$\Phi_\calY:\,\bm{y}'\to\bm{y}$,
\begin{equation*}
    D_\alpha(p_{\bm{y}}:q_{\bm{y}})
    =
    D_\alpha\left(
     p_{\Phi_\calY^{-1}(\bm{y})} :q_{\Phi_\calY^{-1}(\bm{y})}
    \right).
\end{equation*}
This shows
$\calD_{\alpha,f}(\bm{y}) = \calD_{\alpha,f\circ{}\Phi_\calY}(\bm{y})$ everywhere on $\calY$.
In summary, the $\alpha$-discrepancy is invariant to diffeomorphisms on $\calX$ and $\calY$.
\end{proof}

\section{Proof of \Cref{thm:jmj}}

\begin{proof}
We simply denote $J$ instead of $J(\bm{y})$, and $M$
instead of $M(f(\bm{y}))$. One has to remember
that they both depends on $\bm{y}$. By definition,
\begin{equation*}
p_{\bm{y}}(\bar{\bm{y}})
=
\frac{\vert{J}^\T{M}J\vert^{1/2}}{(2\pi)^{d/2} }
\exp\left(
    -\frac{1}{2}
(\bar{\bm{y}}-\bm{y})^\T J^\T M J (\bar{\bm{y}}-\bm{y})\right).
\end{equation*}

If $s_{\bm{y}}(\bar{\bm{y}}) \defeq \exp\left(-\frac{1}{2}\Vert\bar{\bm{y}} - \bm{y}\Vert^2\right)$,
its normalized density is another Gaussian distribution
\begin{equation*}
q_{\bm{y}}(\bar{\bm{y}})
=
\frac{1}{(2\pi)^{d/2}}
\exp\left( -\frac{1}{2}\Vert\bar{\bm{y}} - \bm{y}\Vert^2 \right).
\end{equation*}

Let $\beta=1-\alpha$, then
\begin{align}
&
\int p_{\bm{y}}^{\alpha}( \bar{\bm{y}} ) q_{\bm{y}}^{\beta}( \bar{\bm{y}} ) \dbary
\nonumber\\
=
&
\frac{\vert{J}^\T M J\vert^{\alpha/2}}{(2\pi)^{d/2}}
\int
\exp\bigg(
-\frac{1}{2}
(\bar{\bm{y}}-\bm{y})^\T
\left(\alpha J^\T M J +\beta{I}\right)
(\bar{\bm{y}}-\bm{y}) \bigg)\dbary\nonumber\\
=
&
\frac{\vert{J}^\T M J\vert^{\alpha/2}}
{ \left\vert\alpha J^\T M J +\beta{I} \right\vert^{1/2} }
\int
\frac
{ \left\vert\alpha J^\T M J +\beta{I} \right\vert^{1/2} }
{(2\pi)^{d/2}}
\exp\bigg(
-\frac{1}{2}
(\bar{\bm{y}}-\bm{y})^\T
\left(\alpha J^\T M J +\beta{I}\right)
(\bar{\bm{y}}-\bm{y}) \bigg)\dbary\nonumber\\
=
&
\frac{\vert{J}^\T M J\vert^{\alpha/2}}
{ \left\vert\alpha J^\T M J +\beta{I} \right\vert^{1/2} }.
\end{align}
By \cref{thm:normalize},
\begin{align*}
\calD_{\alpha,f}(\bm{y})
&
\defeq
\inf_{\gamma\in{}C(\calY)} D_\alpha( p_{\bm{y}}:\gamma(\bm{y}) s_{\bm{y}} )\\
&=
\frac{1}{\beta}\left[
    1-
    \left(
\int p_{\bm{y}}^{\alpha}( \bar{\bm{y}} ) q_{\bm{y}}^{\beta}( \bar{\bm{y}} ) \dbary
\right)^{1/\alpha}
\right]
=
\frac{1}{\beta}\left[
    1-
\frac{\vert{J}^\T M J\vert^{1/2}}
{ \left\vert\alpha J^\T M J +\beta{I} \right\vert^{1/2\alpha} }
    \right].
\end{align*}

\end{proof}

Let $\alpha\to0$. The factor $\frac{1}{1-\alpha}\to1$, and
$\alpha$ only appears in the term
\begin{equation*}
    \frac{1}{ \left\vert\alpha J^\T {M} J + (1-\alpha) {I}\right\vert^{1/2\alpha} },
\end{equation*}
which is a ``$1^{\infty}$'' type of limit. We have
\begin{align*}
    \lim_{\alpha\to0}
    \frac{1}{ \left\vert\alpha
       J^\T {M} J + (1-\alpha) {I}\right\vert^{1/2\alpha} }
    =&
    \lim_{\alpha\to0}
    \exp\left(
        -
        \frac{\log \left\vert\alpha J^\T {M} J + (1-\alpha) {I}\right\vert}{2\alpha}
        \right)\\
    =&
    \exp\left(
        -
        \lim_{\alpha\to0}
        \frac{\log \left\vert\alpha J^\T M J + (1-\alpha) {I}\right\vert}{2\alpha}
        \right)\\
    =&
    \exp\left(
        -
        \lim_{\alpha\to0}
        \frac{
            \tr
            \left((\alpha J^\T M J + (1-\alpha) {I})^{-1} (J^\T M J - I) \right)
        }{2}
        \right)\\
    =&
    \exp\left(
        -
        \frac{ \tr \left(J^\T M J - I \right) }{2}
        \right)\\
    =&
    \exp\left(
        -\frac{1}{2} \tr \left(J^\T M J\right)
        +\frac{d}{2}
        \right).
\end{align*}
In summary, we get
\begin{align*}
\calD_{0,f}(\bm{y})
&\defeq
\lim_{\alpha\to0} \calD_{\alpha,f}(\bm{y})
=
1- \left\vert J^\T M J \right\vert^{1/2}
    \exp\left(
        -\frac{1}{2} \tr \left(J^\T M J\right)
        +\frac{d}{2}
        \right)\nonumber\\
&=
1-
    \exp\left(
        \frac{1}{2}
        \log \left\vert J^\T M J \right\vert
        -\frac{1}{2} \tr \left(J^\T M J\right)
        +\frac{d}{2}
        \right).
\end{align*}
Similarly,
\begin{align*}
\calD_{1,f}(\bm{y})
&\defeq
\lim_{\alpha\to1} \calD_{\alpha,f}(\bm{y})\nonumber\\
&=
\frac{1}{-1}
\bigg[
    1-
\lim_{\alpha\to1}
\frac{\vert{J}^\T {M} J \vert^{1/2}}
{ \left\vert\alpha J^\T {M} J + (1-\alpha) {I}\right\vert^{1/2\alpha} }
\bigg(
\frac{1}{2\alpha^2}
\log \left\vert\alpha J^\T {M} J + (1-\alpha) {I}\right\vert\\
&
\quad\quad
- \frac{1}{2\alpha}
\tr\left( (\alpha J^\T {M} J + (1-\alpha) {I})^{-1} (J^\T {M} J - I) \right)
\bigg)
\bigg]\\
&=
\frac{\vert{J}^\T {M} J \vert^{1/2}}{ \left\vert J^\T {M} J \right\vert^{1/2} }
\left(
\frac{1}{2} \log \left\vert J^\T {M} J \right\vert
-\frac{1}{2}
\tr\left( ( J^\T {M} J )^{-1} (J^\T {M} J - I) \right) \right) \\
&=
\frac{1}{2} \log \left\vert J^\T {M} J \right\vert
-\frac{d}{2}
+\frac{1}{2} \tr \left( ( J^\T {M} J )^{-1} \right).
\end{align*}

By assumption $\log p(\bm{y})$ is bounded. Therefore
\begin{align}
\log p(\bm{x},\bm{y})
= \log p(\bm{y}) + \log p(\bm{x}\,\vert\,\bm{y})
= \log p(\bm{y}) -\frac{\lambda}{2} (\bm{x}-f(\bm{y}))^\T M(\bm{x}) (\bm{x}-f(\bm{y}))^\T.
\end{align}
If $\lambda\to\infty$, the second term will dominate.
A Taylor expansion of $f(\bm{y})$ as a linear function of $\bm{y}$ yields the pull-back metric.

\end{document}